\documentclass[11pt,a4paper]{article}
\usepackage[hyperref]{naaclhlt2018}
\usepackage{times}
\usepackage{latexsym}

\usepackage{url}
\usepackage{multirow}
\usepackage{epsfig}
\usepackage{amsmath}
\newcommand*{\affaddr}[1]{#1} 
\newcommand*{\affmark}[1][*]{\textsuperscript{#1}}
\newcommand*{\email}[1]{\texttt{#1}}

\aclfinalcopy 

\title{Query and Output: Generating Words by Querying Distributed Word Representations for Paraphrase Generation}

\author{Shuming Ma\affmark[1], Xu Sun\affmark[1,2], Wei Li\affmark[1], Sujian Li\affmark[1], Wenjie Li\affmark[3], Xuancheng Ren\affmark[1]\\
\affaddr{\affmark[1]MOE Key Lab of Computational Linguistics, School of EECS, Peking University}\\
\affaddr{\affmark[2]Deep Learning Lab, Beijing Institute of Big Data Research, Peking University}\\
\affaddr{\affmark[3]Department of Computing, The Hong Kong Polytechnic University}\\
\email{\{shumingma, xusun, liweitj47, lisujian, renxc\}@pku.edu.cn}\\
\email{cswjli@comp.polyu.edu.hk}
}

\date{}

\begin{document}
\maketitle
\begin{abstract}
  Most recent approaches use the sequence-to-sequence model for paraphrase generation. The existing sequence-to-sequence model tends to memorize the words and the patterns in the training dataset instead of learning the meaning of the words. Therefore, the generated sentences are often grammatically correct but semantically improper. In this work, we introduce a novel model based on the encoder-decoder framework, called Word Embedding Attention Network (WEAN). Our proposed model generates the words by querying distributed word representations (i.e. neural word embeddings), hoping to capturing the meaning of the according words. Following previous work, we evaluate our model on two paraphrase-oriented tasks, namely text simplification and short text abstractive summarization. Experimental results show that our model outperforms the sequence-to-sequence baseline by the BLEU score of 6.3 and 5.5 on two English text simplification datasets, and the ROUGE-2 F1 score of 5.7 on a Chinese summarization dataset. Moreover, our model achieves state-of-the-art performances on these three benchmark datasets.\footnote{The code is available at \url{https://github.com/lancopku/WEAN}} 
\end{abstract}

\section{Introduction}



Paraphrase is a restatement of the meaning of a text using other words. Many natural language generation tasks are paraphrase-orientated, such as text simplification and short text summarization. Text simplification is to make the text easier to read and understand, especially for poor readers, while short text summarization is to generate a brief sentence to describe the short texts (e.g. posts on the social media). Most recent approaches use sequence-to-sequence model for paraphrase generation~\cite{paraphrasegeneration,CaoEA2017}. It compresses the source text information into dense vectors with the neural encoder, and the neural decoder generates the target text using the compressed vectors.


Although neural network models achieve success in paraphrase generation, there are still two major problems. One of the problem is that the existing sequence-to-sequence model tends to memorize the words and the patterns in the training dataset instead of the meaning of the words. The main reason is that the word generator (i.e. the output layer of the decoder) does not model the semantic information. The word generator, which consists of a linear transformation and a softmax operation, converts the Recurrent Neural Network (RNN) output from a small dimension (e.g. 500) to a much larger dimension (e.g. 50,000 words in the vocabulary), where each dimension represents the score of each word. The latent assumption of the word generator is that each word is independent and the score is irrelevant to each other. Therefore, the scores of a word and its synonyms may be of great difference, which means the word generator learns the word itself rather than the relationship between words. 

The other problem is that the word generator has a huge number of parameters. Suppose we have a sequence-to-sequence model with a hidden size of 500 and a vocabulary size of 50,000. The word generator has up to 25 million parameters, which is even larger than other parts of the encoder-decoder model in total. The huge size of parameters will result in slow convergence, because there are a lot of parameters to be learned. Moreover, under the distributed framework, the more parameters a model has, the more bandwidth and memory it consumes. 

To tackle both of the problems, we propose a novel model called Word Embedding Attention Network (WEAN). The word generator of WEAN is attention based, instead of the simple linear softmax operation. In our attention based word generator, the RNN output is a query, the candidate words are the values, and the corresponding word representations are the keys. In order to predict the word, the attention mechanism is used to select the value matching the query most, by means of querying the keys. In this way, our model generates the words according to the distributed word representations (i.e. neural word embeddings) in a retrieval style rather than the traditional generative style. Our model is able to capture the semantic meaning of a word by referring to its embedding. Besides, the attention mechanism has a much smaller number of parameters compared with the linear transformation directly from the RNN output space to the vocabulary space. The reduction of the parameters can increase the convergence rate and speed up the training process. Moreover, the word embedding is updated from three sources: the input of the encoder, the input of the decoder, and the query of the output layer.


Following previous work~\cite{CaoEA2017}, we evaluate our model on two paraphrase-oriented tasks, namely text simplification and short text abstractive summarization. Experimental results show that our model outperforms the sequence-to-sequence baseline by the BLEU score of 6.3 and 5.5 on two English text simplification datasets, and the ROUGE-2 F1 score of 5.7 on a Chinese summarization dataset. Moreover, our model achieves state-of-the-art performances on all of the benchmark datasets.


\section{Proposed Model}

We propose a novel model based on the encoder-decoder framework, which generates the words by querying distributed word representations with the attention mechanism. In this section, we first present the overview of the model architecture. Then, we explain the details of the word generation, especially the way to query word embeddings. 

\subsection{Overview}

\begin{figure*}[tb]
	\centering
	\begin{tabular}{@{}c@{}@{}c@{}@{}c@{}@{}c@{}}
		
		\epsfig{file=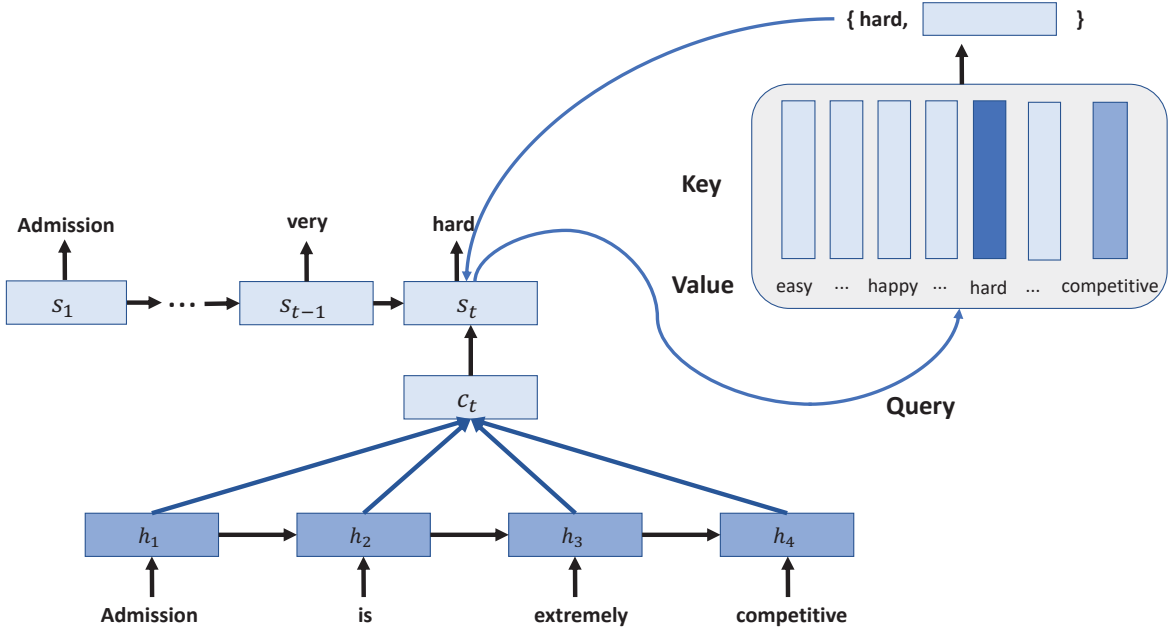,width=1.0\linewidth,clip=}
		
	\end{tabular}
	\caption{An overview of Word Embedding Attention Network.
	}\label{fig1}
	\vspace{-0.1in}
\end{figure*}

Word Embedding Attention Network is based on the encoder-decoder framework, which consists of two components: a source text encoder, and a target text decoder. Figure~\ref{fig1} is an illustration of our model. Given the source texts, the encoder compresses the source texts into dense representation vectors, and the decoder generates the paraphrased texts. To predict a word, the decoder uses the hidden output to query the word embeddings. The word embeddings assess all the candidate words, and return the word whose embedding matches the query most. The selected word is emitted as the predicted token, and its embedding is then used as the input of the LSTM at the next time step. After the back propagation, the word embedding is updated from three sources: the input of the encoder, the input of the decoder, and the query of the output layer. We show the details of our WEAN in the following subsection.

\subsection{Encoder and Decoder}

The goal of the source text encoder is to provide a series of dense representation of complex source texts for the decoder. In our model, the source text encoder is a Long Short-term Memory Network (LSTM), which produces the dense representation $\{h_1,h_2,...,h_N\}$ from the source text $\{x_1,x_2,...,x_N\}$:

The goal of the target text decoder is to generate a series of paraphrased words from the dense representation of source texts. Fisrt, the LSTM of the decoder compute the dense representation of generated words $s_t$. Then, the dense representations are fed into an attention layer~\cite{attention} to generate the context vector $c_t$, which captures context information of source texts. Attention vector $c_t$ is calculated by the weighted sum of encoder hidden states:
\begin{equation}\label{attention1}
c_t=\sum_{i=1}^{N}{\alpha_{ti}h_{i}}
\end{equation}
\begin{equation}\label{attention2}
\alpha_{ti}=\frac{e^{g(s_{t},h_{i})}}{\sum_{j=1}^{N}{e^{g(s_{t},h_{j})}}}
\end{equation}
where $g(s_{t},h_{i})$ is an attentive score between the decoder hidden state $s_t$ and the encoder hidden state $h_i$.

In this way, $c_t$ and $s_t$ respectively represent the context information of source texts and the target texts at the $t^{th}$ time step.

\subsection{Word Generation by Querying Word Embedding}\label{sec_query}

For the current sequence-to-sequence model, the word generator computes the distribution of output words $y_t$ in a generative style:
\begin{equation}\label{generator2}
p(y_t)=softmax{(Ws_t)}
\end{equation}
where $W \in R^{k \times V}$ is a trainable parameter matrix, $k$ is hidden size, and $V$ is the number of words in the vocabulary. When the vocabulary is large, the number of parameters will be huge.

Our model generates the words in a retrieval style rather than the traditional generative style, by querying the word embeddings. We denote the combination of the source context vector $c_t$ and the target context vector $s_t$ as the query $q_t$:
\begin{equation}
q_t=\tanh(W_c[s_t;c_t])
\end{equation}
The candidate words $w_i$ and their corresponding embeddings $e_i$ are paired as the key-value pairs $\{w_i, e_i\} (i=1,2,...,n)$, where $n$ is the number of candidate words. We give the details of how to determine the set of candidate words in Section~\ref{selection}. Our model uses $q_t$ to query the key-value pairs $\{w_i, e_i\} (i=1,2,...,n)$ by evaluating the relevance between the query $q_t$ and each word vector $e_i$ with a score function $f(q_t, e_i)$. The query process can be regarded as the attentive selection of the word embeddings. We borrow the attention energy functions~\cite{stanfordattention} as the relevance score function $f(q_t, e_i)$:
\begin{eqnarray}\label{query_attn}
f(q_t, e_i)=
\begin{cases}
q_{t}^{T}e_i & \text{dot}\\
q_t^TW_ae_i & \text{general}\\
v^T\tanh(W_qq_t+W_ee_i) & \text{concat}
\end{cases}
\end{eqnarray}
where $W_q$ and $W_e$ are two trainable parameter matrices, and $v^T$ is a trainable parameter vector. In implementation, we select the general attention function as the relevance score function, based on the performance on the validation sets. The key-value pair with the highest score $\{w_t, e_t\}$ is selected. At the test stage, the decoder generates the key $w_t$ as the $t^{th}$ predicted word, and inputs the value $e_t$ to the LSTM unit at the $t+1^{th}$ time step. At the training stage, the scores are normalized as the word probability distribution:
\begin{equation}
p(y_t)=softmax{(f(q_t, e_i))}
\end{equation}

\subsection{Selection of Candidate Key-value Pairs}\label{selection}

As described in Section~\ref{sec_query}, the model generates the words in a retrieval style, which selects a word according to its embedding from a set of candidate key-value pairs. We now give the details of how to obtain the set of candidate key-value pairs. We extract the vocabulary from the source text in the training set, and select the $n$ most frequent words as the candidate words. We reuse the embeddings of the decoder inputs as the values of the candidate words, which means that the decoder input and the predicted output share the same vocabulary and word embeddings. Besides, we do not use any pretrained word embeddings in our model, so that all of the parameters are learned from scratch.



\subsection{Training}

Although our generator is a retrieval style, WEAN is as differentiable as the sequence-to-sequence model. The objective of training is to minimize the cross entropy between the predicted word probability distribution and the golden one-hot distribution:
\begin{equation}
L=-\sum_i{\hat{y_i}\log{p(y_i)}}
\end{equation}
We use Adam optimization method to train the model, with the default hyper-parameters: the learning rate $\alpha=0.001$, and $\beta_{1}=0.9$, $\beta_{2}=0.999$, $\epsilon=1e-8$.


\section{Experiments}

Following the previous work~\cite{CaoEA2017}, we test our model on the following two paraphrase orientated tasks: text simplification and short text abstractive summarization.

\subsection{Text Simplification}

\subsubsection{Datasets}

The datasets are both from the alignments between English Wikipedia website\footnote{http://en.wikipedia.org} and Simple English Wikipedia website.\footnote{http://simple.wikipedia.org} The Simple English Wikipedia is built for ``the children and adults who are learning the English language'', and the articles are composed with ``easy words and short sentences''. Therefore, Simple English Wikipedia is a natural public simplified text corpus. 

\begin{itemize}
\item \textbf{Parallel Wikipedia Simplification Corpus (PWKP).} PWKP~\cite{ZhuEA2010} is a widely used benchmark for evaluating text simplification systems. It consists of aligned complex text from English WikiPedia (as of Aug. 22nd, 2009) and simple text from Simple Wikipedia (as of Aug. 17th, 2009). The dataset contains 108,016 sentence pairs, with 25.01 words on average per complex sentence and 20.87 words per simple sentence. Following the previous work~\cite{ZhangEA2017}, we remove the duplicate sentence pairs, and split the corpus with 89,042 pairs for training, 205 pairs for validation and 100 pairs for test.

\item\textbf{English Wikipedia and Simple English Wikipedia (EW-SEW).} EW-SEW is a publicly available dataset provided by Hwang et al.~\shortcite{HwangEA2015}. To build the corpus, they first align the complex-simple sentence pairs, score the semantic similarity between the complex sentence and the simple sentence, and classify each sentence pair as a good, good partial, partial, or bad match. Following the previous work~\cite{NisioiEA2017}, we discard the unclassified matches, and use the good matches and partial matches with a scaled threshold greater than 0.45. The corpus contains about 150K good matches and 130K good partial matches. We use this corpus as the training set, and the dataset provided by Xu et al.~\cite{XuEA2016} as the validation set and the test set. The validation set consists of 2,000 sentence pairs, and the test set contains 359 sentence pairs. Besides, each complex sentence is paired with 8 reference simplified sentences provided by Amazon Mechanical Turk workers.
\end{itemize}

\subsubsection{Evaluation Metrics}

Following the previous work~\cite{NisioiEA2017,lcsts}, we evaluate our model with different metrics on two tasks.

\begin{itemize}
\item \noindent\textbf{Automatic evaluation.} We use the BLEU score~\cite{bleu} as the automatic evaluation metric. BLEU is a widely used metric for machine translation and text simplification, which measures the agreement between the model outputs and the gold references. The references can be either single or multiple. In our experiments, the references are single on PWKP, and multiple on EW-SEW.

\item \noindent\textbf{Human evaluation.} Human evaluation is essential to evaluate the quality of the model outputs. Following Nisioi et al.~\shortcite{NisioiEA2017} and Zhang et al.~\shortcite{ZhangEA2017}, we ask the human raters to rate the simplified text in three dimensions: Fluency, Adequacy and Simplicity. Fluency assesses whether the outputs are grammatically right and well formed. Adequacy represents the meaning preservation of the simplified text. Both the scores of fluency and adequacy range from 1 to 5 (1 is very bad and 5 is very good). Simplicity shows how simpler the model outputs are than the source text, which ranges from 1 to 5.

\end{itemize}

\subsubsection{Settings}

Our proposed model is based on the encoder-decoder framework. The encoder is implemented on LSTM, and the decoder is based on LSTM with Luong style attention~\cite{stanfordattention}. We tune our hyper-parameter on the development set. The model has two LSTM layers. The hidden size of LSTM is 256, and the embedding size is 256. We use Adam optimizer~\cite{KingmaBa2014} to learn the parameters, and the batch size is set to be 64. We set the dropout rate~\cite{dropout} to be 0.4. All of the gradients are clipped when the norm exceeds 5. 


\begin{table}[tb]
	\centering
	\begin{tabular}{lc}
		\hline
		PWKP & BLEU \\
		\hline
		PBMT~\cite{WubbenEA2012} & 46.31 \\
		Hybrid~\cite{NarayanEA2014} & 53.94 \\
		EncDecA~\cite{ZhangEA2017} & 47.93 \\
		DRESS~\cite{ZhangEA2017} &  34.53  \\
		DRESS-LS~\cite{ZhangEA2017} &  36.32   \\ \hline
		Seq2seq (our implementation) & 48.26 \\
		\textbf{WEAN (our proposal)} &  \textbf{54.54}  \\ 
		\hline
	\end{tabular}
	\caption{Automatic evaluation of our model and other related systems on PWKP datasets. The results are reported on the test sets.}
	\label{tab_PWKP_bleu}
\end{table}

\begin{table}[tb]
	\centering
	\begin{tabular}{lc}
		\hline
		EW-SEW & BLEU  \\
		\hline
		PBMT-R~\cite{WubbenEA2012} &  67.79  \\
		Hybrid~\cite{NarayanEA2014} & 48.97 \\
		SBMT-SARI~\cite{XuEA2016} &  73.62   \\ 
		NTS~\cite{NisioiEA2017} &  84.70  \\
		NTS-w2v~\cite{NisioiEA2017} & 87.50  \\
		EncDecA~\cite{ZhangEA2017} & 88.85 \\
		DRESS~\cite{ZhangEA2017} &  77.18   \\
		DRESS-LS~\cite{ZhangEA2017} &  80.12   \\ \hline
		Seq2seq (our implementation) & 88.97 \\  
		\textbf{WEAN (our proposal)} &  \textbf{94.45}    \\ 
		\hline
	\end{tabular}
	\caption{Automatic evaluation of our model and other related systems on EW-SEW datasets. The results are reported on the test sets.}
	\label{tab_EWSEW_bleu}
\end{table}

\begin{table}[tb]
	\centering
	\begin{tabular}{@{~} l@{~} @{~}c@{~} c@{~} c@{~} c@{~}}
		\hline
		PWKP & Fluency & Adequacy & Simplicity & All \\
		\hline
		NTS-w2v & 3.54 & 3.47 & 3.38  & 3.46\\
		DRESS-LS & 3.68 & 3.55 & 3.50  & 3.58 \\ 
		WEAN  & \textbf{3.77} & \textbf{3.66} & \textbf{3.58} & \textbf{3.67} \\ \hline
		Reference & 3.76 & 3.60 & 3.44 & 3.60 \\ \hline
		\multicolumn{5}{c}{} \\
		\hline
		EW-SEW & Fluency & Adequacy & Simplicity & All \\
		\hline
		PBMT-R & 3.36 & 2.92 & 3.37 & 3.22 \\
		SBMT-SARI & 3.41 & \textbf{3.63} & 3.25 & 3.43 \\ 
		NTS-w2v & 3.56 & 3.52 &  3.42 & 3.50 \\
		DRESS-LS & 3.59 & 3.43 &  \textbf{3.65} & 3.56\\
		WEAN & \textbf{3.61} &  3.56 & \textbf{3.65}  & \textbf{3.61} \\ \hline
		Reference  & 3.71 &  3.64 &  3.45 & 3.60 \\ \hline
	\end{tabular}
	\caption{Human evaluation of our model and other related systems on PWKP and EW-SEW datasets. The results are reported on the test sets.}
	\label{tab_human}
\end{table}

\subsubsection{Baselines}

We compare our model with several neural text simplification systems. 
\begin{itemize}

\item \textbf{Seq2seq} is our implementation of the sequence-to-sequence model with attention mechanism, which is the most popular neural model for text generation. 
\item \textbf{NTS} and \textbf{NTS-w2v}~\cite{NisioiEA2017} are two sequence-to-sequence model with extra mechanism like prediction ranking, and NTS-w2v uses a pretrain word2vec. 
\item \textbf{DRESS} and \textbf{DRESS-LS}~\cite{ZhangEA2017} are two deep reinforcement learning sentence simplification models. 
\item \textbf{EncDecA} is a model based on the encoder-decoder with attention, implemented by \citet{ZhangEA2017}. 
\item \textbf{PBMT-R}~\cite{WubbenEA2012} is a phrase based machine translation model which reranks the outputs. \item \textbf{Hybrid}~\cite{NarayanEA2014} is a hybrid approach which combines deep semantics and mono-lingual machine translation. 
\item \textbf{SBMT-SARI}~\cite{XuEA2016} is a syntax-based machine translation model which is trained on PPDB dataset~\cite{PPDB} and tuned with SARI.

\end{itemize}

\subsubsection{Results}

We compare WEAN with state-of-the-art models for text simplification. Table~\ref{tab_PWKP_bleu} and Table~\ref{tab_EWSEW_bleu} summarize the results of the automatic evaluation. On PWKP dataset, we compare WEAN with PBMT, Hybrid, EncDecA, DRESS and DRESS-LS. WEAN achieves a BLEU score of 54.54, outperforming all of the previous systems. On EW-SEW dataset, we compare WEAN with PBMT-R, Hybrid, SBMT-SARI, and the neural models described above. We do not find any public release code of PBMT-R and SBMT-SARI. Fortunately, Xu et al.~\shortcite{XuEA2016} provides the predictions of PBMT-R and SBMT-SARI on EW-SEW test set, so that we can compare our model with these systems. It shows that the neural models have better performance in BLEU, and WEAN achieves the best BLEU score with 94.45.

We perform the human evaluation of WEAN and other related systems, and the results are shown in Table~\ref{tab_human}. DRESS-LS is based on the reinforcement learning, and it encourages the fluency, simplicity and relevance of the outputs. Therefore, it achieves a high score in our human evaluation. WEAN gains a even better score than DRESS-LS. Besides, WEAN generates more adequate and simpler outputs than the reference on PWKP. The predictions of SBMT-SARI are the most adequate among the compared systems on EW-SEW. In general, WEAN outperforms all of the other systems, considering the balance of fluency, adequate and simplicity. We conduct significance tests based on t-test. The significance tests suggest that WEAN has a very significant improvement over baseline, with $p \leq 0.001$ over DRESS-LS in all of the dimension on PWKP, $p \leq 0.05$ over DRESS-LS in the dimension of fluency, $p \leq 0.005$ over NTS-w2v in the dimension of simplicity and $p \leq 0.005$ over DRESS-LS in the dimension of all.

\begin{table}[tb]
	\centering
	\begin{tabular}{@{}l@{}c@{}c@{}c@{}}
		\hline
		\multicolumn{1}{l}{\textbf{LCSTS}} &
		\multicolumn{1}{c}{\textbf{R-1}} & 
		\multicolumn{1}{c}{\textbf{R-2}} &  
		\multicolumn{1}{c}{\textbf{R-L}}   \\ \hline
		RNN-W\cite{lcsts} &  17.7  & 8.5  &  15.8 \\
		RNN\cite{lcsts} & 21.5 & 8.9 & 18.6 \\
		RNN-cont-W\cite{lcsts} & 26.8  & 16.1  &  24.1  \\
		RNN-cont\cite{lcsts} & 29.9 & 17.4 & 27.2 \\
		SRB\cite{srb} & 33.3 & 20.0 & 30.1 \\
		CopyNet-W\cite{copynet} & 35.0 & 22.3 & 32.0 \\ 
		CopyNet\cite{copynet} & 34.4 & 21.6  &  31.3 \\
		RNN-dist\cite{distraction} & 35.2 & 22.6 & 32.5 \\
		DRGD\cite{DRGD} & 37.0 & 24.2 & 34.2  \\ 
		\hline
		Seq2seq & 32.1 & 19.9 & 29.2 \\
		\textbf{WEAN} & \textbf{37.8} & \textbf{25.6} & \textbf{35.2} \\
		\hline
		
	\end{tabular}
	\caption{ROUGE F1 score on the LCSTS test set. R-1, R-2, and R-L denote ROUGE-1, ROUGE-2, and ROUGE-L, respectively. The models with a suffix of `W' in the table are word-based, while the rest of models are character-based.}\label{tab_lcsts}
\end{table}

\subsection{Large Scale Text Summarization}


\subsubsection{Dataset}

\noindent\textbf{Large Scale Chinese Social Media Short Text Summarization Dataset (LCSTS):} LCSTS is constructed by \citet{lcsts}. The dataset consists of more than 2,400,000 text-summary pairs, constructed from a famous Chinese social media website called Sina Weibo.\footnote{\url{http://weibo.com}} It is split into three parts, with 2,400,591 pairs in PART I, 10,666 pairs in PART II and 1,106 pairs in PART III. All the text-summary pairs in PART II and PART III are manually annotated with relevant scores ranged from 1 to 5. We only reserve pairs with scores no less than 3, leaving 8,685 pairs in PART II and 725 pairs in PART III. Following the previous work~\cite{lcsts}, we use PART I as training set, PART II as validation set, and PART III as test set. 

\subsubsection{Evaluation Metrics}

Our evaluation metric is ROUGE score~\cite{rouge}, which is popular for summarization evaluation. The metrics compare an automatically produced summary against the reference summaries, by computing overlapping lexical units, including unigram, bigram, trigram, and longest common subsequence (LCS). Following previous work~\cite{abs,lcsts}, we use ROUGE-1 (unigram), ROUGE-2 (bi-gram) and ROUGE-L (LCS) as the evaluation metrics in the reported experimental results.

\subsubsection{Settings}

The vocabularies are extracted from the training sets, and the source contents and the summaries share the same vocabularies. We tune the hyper-parameters based on the ROUGE scores on the validation sets.
In order to alleviate the risk of word segmentation mistakes, we split the Chinese sentences into characters. We prune the vocabulary size to 4,000, which covers most of the common characters.
We set the word embedding size and the hidden size to 512, the number of LSTM layers of the encoder is 2, and the number of LSTM layers of the decoder is 1. The batch size is 64, and we do not use dropout~\cite{dropout} on this dataset. Following the previous work~\cite{DRGD}, we implement a beam search optimization, and set the beam size to 5. 

\subsubsection{Baselines}

We compare our model with the state-of-the-art baselines.

\begin{itemize}
\item \textbf{RNN} and \textbf{RNN-cont} are two sequence-to-sequence baseline with GRU encoder and decoder, provided by~\citet{lcsts}. 
\item \textbf{RNN-dist}~\cite{distraction} is a distraction-based neural model, which the attention mechanism focuses on the different parts of the source content. 
\item \textbf{CopyNet}~\cite{copynet} incorporates a copy mechanism to allow part of the generated summary is copied from the source content. 
\item \textbf{SRB}~\cite{srb} is a sequence-to-sequence based neural model with improving the semantic relevance between the input text and the output summary. 
\item \textbf{DRGD}~\cite{DRGD} is a deep recurrent generative decoder model, combining the decoder with a variational autoencoder. 
\item \textbf{Seq2seq} is our implementation of the sequence-to-sequence model with the attention mechanism. 
\end{itemize}

\subsubsection{Results}

We report the ROUGE F1 score of our model and the baseline models on the test sets.
Table~\ref{tab_lcsts} summarizes the comparison between our model and the baselines. Our model achieves the score of 37.8 ROUGE-1, 25.6 ROUGE-2, and 35.2 ROUGE-L, outperforming all of the previous models. First, we compare our model with the sequence-to-sequence model. It shows that our model significant outperforms the sequence-to-sequence baseline with a large margin of 5.7 ROUGE-1, 5.7 ROUGE-2, and 6.0 ROUGE-L. Then, we compare our model with other related models. The state-of-the-art model is DRGD~\cite{DRGD}, which obtains the score of 37.0 ROUGE-1, 24.2 ROUGE-2, and 34.2 ROUGE-L. Our model has a relative gain of 0.8 ROUGE-1, 1.4 ROUGE-2 and 1.0 ROUGE-L over the state-of-the-art models.

\begin{table}[tb]
	\centering
	\begin{tabular}{c|ccc}
		\hline
		\multicolumn{1}{c|}{\textbf{\#Param}} &
		\multicolumn{1}{c}{\textbf{PWKP}} &
		\multicolumn{1}{c}{\textbf{EWSEW}} &
		\multicolumn{1}{c}{\textbf{LCSTS}}   \\ \hline
		Seq2seq & 12.80M & 12.80M & 2.05M \\
		WEAN & 0.13M & 0.13M & 0.52M \\
		\hline
	\end{tabular}
	\caption{The number of the parameters in the output layer. The numbers of rest parameters between Seq2seq and WEAN are the same.}\label{tab_param}
\end{table}

\begin{figure}[tb]
	\centering
	\begin{tabular}{@{}c@{}@{}c@{}@{}c@{}@{}c@{}}
		
		\epsfig{file=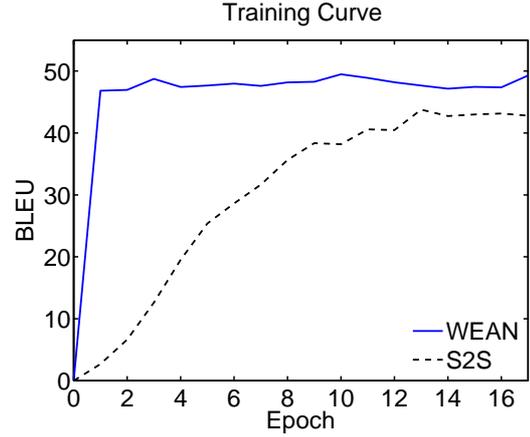,width=0.9\linewidth,clip=} \\

	\end{tabular}
	\caption{The training curve of WEAN and Seq2seq on the PWKP validation set.
	}\label{fig2}
	\vspace{-0.1in}
\end{figure}

\section{Analysis and Discussion}

\subsection{Reducing Parameters}

Our WEAN reduces a large number of the parameters in the output layer. To analyze the parameter reduction, we compare our WEAN model with the sequence-to-sequence model. Table~\ref{tab_param} lists the number of  the parameters in the output layers of two models. Both PWKP and EWSEW　have the vocabulary size of 50000 words and the hidden size of 256, resulting $50000 \times 256 = 12,800,000$ parameters. LCSTS has a vocabulary size of 4000 and the hidden size of 512, so the seq2seq has $4000 \times 512 = 2,048,000$ parameters in the output layers. WEAN only has two parameter matrices and one parameter vector at most in Equation~\ref{query_attn}, without regard to the vocabulary size. It has $256 \times 256 \times 2 + 256 = 131,328$ parameters on PWKP and EWSEW, and $512 \times 512 \times 2 + 512 = 524,800$ parameters on LCSTS. Besides, WEAN does not have any extra parameters in the other part of the model.

\subsection{Speeding up Convergence}

Figure~\ref{fig2} shows the training curve of WEAN and Seq2seq on the PWKP validation set. WEAN achieve near the optimal score in only 2-3 epochs, while Seq2seq takes more than 15 epochs to achieve the optimal score. Therefore, WEAN has much faster convergence rate, compared with Seq2seq. With the much faster training speed, WEAN does not suffer loss in BLEU, and even improve the BLEU score.

\begin{table*}[tb]
	\centering
	\begin{tabular}{| l  p{13.3cm}@{~} |}
		\hline
		Source & \textsl{Yoghurt or yogurt is a dairy product produced by bacterial fermentation of milk .} \\
		Reference & \textsl{Yoghurt or yogurt is a dairy product \textbf{made} by bacterial fermentation of milk .} \\
		NTS & \textsl{\textbf{. or yoghurt} is a dairy product \textbf{produced} by bacterial fermentation of milk .} \\
		NTS-w2v & \textsl{\textbf{It is made} by bacterial fermentation of milk .} \\
		PBMT-R & \textsl{Yoghurt or yogurt is a dairy product \textbf{produced} by bacterial fermentation \textbf{of .}}\\
		SBMT-SARI & \textsl{Yogurt or yogurt is a dairy product \textbf{drawn up} by bacterial fermentation of milk .}\\
		WEAN & \textsl{Yoghurt or yogurt is a dairy product \textbf{made} by bacterial fermentation of milk .} \\
		\hline
		\hline
		Source & \textsl{Depending on the context, another closely-related meaning of constituent is that of a citizen residing in the area governed, represented, or otherwise served by a politician; sometimes this is restricted to citizens who elected the politician.} \\ 
		Reference & \textsl{\textbf{The word constituent can also be used to refer to} a citizen \textbf{who lives} in the area that is governed, represented, or otherwise served by a politician; sometimes \textbf{the word} is restricted to citizens who elected the politician.} \\
		NTS & \textsl{Depending on the context, another closely-related meaning of constituent is that of a citizen \textbf{living} in the area governed, represented, or otherwise served by a politician; sometimes this is restricted to citizens who elected the politician.} \\
		NTS-w2v & \textsl{This is restricted to citizens who elected the politician.} \\
		PBMT-R & \textsl{Depending on the context and meaning of closely-related \textbf{siemens-martin -rrb- is a} citizen living in the area, or otherwise, \textbf{was governed by a 1924-1930 shurba}; this is restricted to people who elected \textbf{it}.} \\
		SBMT-SARI & \textsl{\textbf{In terms of} the context, another closely-related \textbf{sense of the component} is that of a citizen \textbf{living} in the area \textbf{covered, make up, or if not, served by a policy}; sometimes this is \textbf{limited} to the people who elected the \textbf{policy}.}
		\\
		WEAN & \textsl{Depending on the context, another closely-related meaning of constituent is that of a citizen \textbf{who lives} in the area governed, represented, or otherwise served by a politician; sometimes \textbf{the word} is restricted to citizens who elected the politician.} \\
		\hline
	\end{tabular}
	\caption{Two examples of different text simplification system outputs in EW-SEW dataset. Differences from the source texts are shown in bold.
	}\label{tab_case}
	\vspace{-0.1in}
\end{table*}

\subsection{Case Study}

Table~\ref{tab_case} shows two examples of different text simplification system outputs on EW-SEW. For the first example, NTS, NTS-w2v and PBMT-R miss some essential constituents, so that the sentences are incomplete and not fluent. SBMT-SARI generates a fluent sentence, but the output does not preserve the original meaning. The predicted sentence of WEAN is fluent, simple, and the same as the reference. For the second example, NTS-w2v omits so many words that it lacks a lot of information. PBMT-R generates some irrelevant words, like 'siemens-martin', '-rrb-', and '-shurba', which hurts the fluency and adequacy of the generated sentence. SBMT-SARI is able to generate a fluent sentence, but the meaning is different from the source text, and even more difficult to understand.
Compared with the statistic model, WEAN generates a more fluent sentence. Besides, WEAN can capture the semantic meaning of the word by querying the word embeddings, so the generated sentence is semantically correct, and very close to the original meaning.

\section{Related Work}

Our work is related to the encoder-decoder framework~\cite{ChoEA2014} and the attention mechanism~\cite{attention}. Encoder-decoder framework, like sequence-to-sequence model, has achieved success in machine translation~\cite{seq2seq,mapattention,stanfordattention,LinEA2018}, text summarization~\cite{abs,ras,ibmsummarization,randomwalk,MaSun2017}, and other natural language processing tasks~\cite{table2text}. There are many other methods to improve neural attention model~\cite{mapattention,stanfordattention}.

~\citet{ZhuEA2010} constructs a wikipedia dataset, and proposes a tree-based simplification model. \citet{Woodsend2011} introduces a data-driven model based on quasi-synchronous grammar, which captures structural mismatches and complex rewrite operations. \citet{WubbenEA2012} presents a method for text simplification using phrase based machine translation with re-ranking the outputs. \citet{Kauchak2013} proposes a text simplification corpus, and evaluates language modeling for text simplification on the proposed corpus. \citet{NarayanEA2014} propose a hybrid approach to sentence simplification which combines deep semantics and monolingual machine translation. \citet{HwangEA2015} introduces a parallel simplification corpus by evaluating the similarity between the source text and the simplified text based on WordNet. \citet{LigthLS} propose an unsupervised approach to lexical simplification that makes use of word vectors and require only regular corpora. \citet{XuEA2016} design automatic metrics for text simplification. Recently, most works focus on the neural sequence-to-sequence model. \citet{NisioiEA2017} present a sequence-to-sequence model, and re-ranks the predictions with BLEU and SARI. \citet{ZhangEA2017} propose a deep reinforcement learning model to improve the simplicity, fluency and adequacy of the simplified texts. \citet{CaoEA2017} introduce a novel sequence-to-sequence model to join copying and restricted generation for text simplification.

\citet{abs} first used an attention-based encoder to compress texts and a neural network language decoder to generate summaries.
Following this work, recurrent encoder was introduced to text summarization, and gained better performance~\cite{rnnheadline,ras}. Towards Chinese texts, \citet{lcsts}
built a large corpus of Chinese short text summarization. To deal with unknown word problem, \citet{ibmsummarization} proposed a generator-pointer model
so that the decoder is able to generate words in source texts. \citet{copynet} also solved this issue by incorporating copying mechanism.

\section{Conclusion}

We propose a novel model based on the encoder-decoder framework, which generates the words by querying distributed word representations. Experimental results show that our model outperforms the sequence-to-sequence baseline by the BLEU score of 6.3 and 5.5 on two English text simplification datasets, and the ROUGE-2 F1 score of 5.7 on a Chinese summarization dataset. Moreover, our model achieves state-of-the-art performances on these three benchmark datasets.

\section*{Acknowledgements}

This work was supported in part by National Natural Science Foundation of China (No. 61673028), National High Technology Research and Development Program of China (863 Program, No. 2015AA015404), and the National Thousand Young Talents Program. Xu Sun is the corresponding author of this paper.

\nocite{SunEA2017,SunWei2017,dnerre,mesimp,amr,discourse}

\bibliographystyle{acl_natbib}
\bibliography{wean}

\begin{thebibliography}{}
\expandafter\ifx\csname natexlab\endcsname\relax\def\natexlab#1{#1}\fi

\bibitem[{Bahdanau et~al.(2014)Bahdanau, Cho, and Bengio}]{attention}
Dzmitry Bahdanau, Kyunghyun Cho, and Yoshua Bengio. 2014.
\newblock Neural machine translation by jointly learning to align and
  translate.
\newblock {\em CoRR\/} abs/1409.0473.

\bibitem[{Cao et~al.(2017)Cao, Luo, Li, and Li}]{CaoEA2017}
Ziqiang Cao, Chuwei Luo, Wenjie Li, and Sujian Li. 2017.
\newblock Joint copying and restricted generation for paraphrase.
\newblock In {\em Proceedings of the Thirty-First {AAAI} Conference on
  Artificial Intelligence\/}. pages 3152--3158.

\bibitem[{Chen et~al.(2016)Chen, Zhu, Ling, Wei, and Jiang}]{distraction}
Qian Chen, Xiaodan Zhu, Zhenhua Ling, Si~Wei, and Hui Jiang. 2016.
\newblock Distraction-based neural networks for modeling documents.
\newblock In {\em Proceedings of the 25th International Joint Conference on
  Artificial Intelligence (IJCAI 2015)\/}. AAAI, New York, NY.

\bibitem[{Cheng and Lapata(2016)}]{discourse}
Jianpeng Cheng and Mirella Lapata. 2016.
\newblock Neural summarization by extracting sentences and words.
\newblock In {\em Proceedings of the 54th Annual Meeting of the Association for
  Computational Linguistics, {ACL} 2016, August 7-12, 2016, Berlin, Germany,
  Volume 1: Long Papers\/}.

\bibitem[{Cho et~al.(2014)Cho, van Merrienboer, G{\"{u}}l{\c{c}}ehre, Bahdanau,
  Bougares, Schwenk, and Bengio}]{ChoEA2014}
Kyunghyun Cho, Bart van Merrienboer, {\c{C}}aglar G{\"{u}}l{\c{c}}ehre, Dzmitry
  Bahdanau, Fethi Bougares, Holger Schwenk, and Yoshua Bengio. 2014.
\newblock Learning phrase representations using {RNN} encoder-decoder for
  statistical machine translation.
\newblock In {\em Proceedings of the 2014 Conference on Empirical Methods in
  Natural Language Processing, {EMNLP} 2014\/}. pages 1724--1734.

\bibitem[{Chopra et~al.(2016)Chopra, Auli, and Rush}]{ras}
Sumit Chopra, Michael Auli, and Alexander~M. Rush. 2016.
\newblock Abstractive sentence summarization with attentive recurrent neural
  networks.
\newblock In {\em {NAACL} {HLT} 2016, The 2016 Conference of the North American
  Chapter of the Association for Computational Linguistics: Human Language
  Technologies\/}. pages 93--98.

\bibitem[{Ganitkevitch et~al.(2013)Ganitkevitch, Durme, and
  Callison{-}Burch}]{PPDB}
Juri Ganitkevitch, Benjamin~Van Durme, and Chris Callison{-}Burch. 2013.
\newblock {PPDB:} the paraphrase database.
\newblock In {\em Human Language Technologies: Conference of the North American
  Chapter of the Association of Computational Linguistics, Proceedings\/}.
  pages 758--764.

\bibitem[{Glava\v{s} and \v{S}tajner(2015)}]{LigthLS}
Goran Glava\v{s} and Sanja \v{S}tajner. 2015.
\newblock Simplifying lexical simplification: Do we need simplified corpora?
\newblock In {\em Proceedings of the 53rd Annual Meeting of the Association for
  Computational Linguistics, {ACL}\/}. pages 63--68.

\bibitem[{Gu et~al.(2016)Gu, Lu, Li, and Li}]{copynet}
Jiatao Gu, Zhengdong Lu, Hang Li, and Victor O.~K. Li. 2016.
\newblock Incorporating copying mechanism in sequence-to-sequence learning.
\newblock In {\em Proceedings of the 54th Annual Meeting of the Association for
  Computational Linguistics, {ACL} 2016\/}.

\bibitem[{Hu et~al.(2015)Hu, Chen, and Zhu}]{lcsts}
Baotian Hu, Qingcai Chen, and Fangze Zhu. 2015.
\newblock {LCSTS:} {A} large scale chinese short text summarization dataset.
\newblock In {\em Proceedings of the 2015 Conference on Empirical Methods in
  Natural Language Processing, {EMNLP} 2015, Lisbon, Portugal, September 17-21,
  2015\/}. pages 1967--1972.

\bibitem[{Hwang et~al.(2015)Hwang, Hajishirzi, Ostendorf, and Wu}]{HwangEA2015}
William Hwang, Hannaneh Hajishirzi, Mari Ostendorf, and Wei Wu. 2015.
\newblock Aligning sentences from standard wikipedia to simple wikipedia.
\newblock In {\em {NAACL} {HLT} 2015\/}. pages 211--217.

\bibitem[{Jean et~al.(2015)Jean, Cho, Memisevic, and Bengio}]{mapattention}
S{\'{e}}bastien Jean, KyungHyun Cho, Roland Memisevic, and Yoshua Bengio. 2015.
\newblock On using very large target vocabulary for neural machine translation.
\newblock In {\em Proceedings of the 53rd Annual Meeting of the Association for
  Computational Linguistics, {ACL} 2015\/}. pages 1--10.

\bibitem[{Kauchak(2013)}]{Kauchak2013}
David Kauchak. 2013.
\newblock Improving text simplification language modeling using unsimplified
  text data.
\newblock In {\em Proceedings of the 51st Annual Meeting of the Association for
  Computational Linguistics, {ACL}\/}. pages 1537--1546.

\bibitem[{Kingma and Ba(2014)}]{KingmaBa2014}
Diederik~P. Kingma and Jimmy Ba. 2014.
\newblock Adam: {A} method for stochastic optimization.
\newblock {\em CoRR\/} abs/1412.6980.

\bibitem[{Li et~al.(2017)Li, Lam, Bing, and Wang}]{DRGD}
Piji Li, Wai Lam, Lidong Bing, and Zihao Wang. 2017.
\newblock Deep recurrent generative decoder for abstractive text summarization.
\newblock In {\em Proceedings of the 2017 Conference on Empirical Methods in
  Natural Language Processing, {EMNLP} 2017, Copenhagen, Denmark, September
  9-11, 2017\/}. pages 2091--2100.

\bibitem[{Lin and Hovy(2003)}]{rouge}
Chin{-}Yew Lin and Eduard~H. Hovy. 2003.
\newblock Automatic evaluation of summaries using n-gram co-occurrence
  statistics.
\newblock In {\em Human Language Technology Conference of the North American
  Chapter of the Association for Computational Linguistics, {HLT-NAACL}
  2003\/}.

\bibitem[{Lin et~al.(2018)Lin, Ma, Su, and Sun}]{LinEA2018}
Junyang Lin, Shuming Ma, Qi~Su, and Xu~Sun. 2018.
\newblock Decoding-history-based adaptive control of attention for neural
  machine translation.
\newblock {\em CoRR\/} abs/1802.01812.

\bibitem[{Liu et~al.(2017)Liu, Wang, Sha, Chang, and Sui}]{table2text}
Tianyu Liu, Kexiang Wang, Lei Sha, Baobao Chang, and Zhifang Sui. 2017.
\newblock Table-to-text generation by structure-aware seq2seq learning.
\newblock {\em CoRR\/} abs/1711.09724.

\bibitem[{Lopyrev(2015)}]{rnnheadline}
Konstantin Lopyrev. 2015.
\newblock Generating news headlines with recurrent neural networks.
\newblock {\em CoRR\/} abs/1512.01712.

\bibitem[{Luong et~al.(2015)Luong, Pham, and Manning}]{stanfordattention}
Thang Luong, Hieu Pham, and Christopher~D. Manning. 2015.
\newblock Effective approaches to attention-based neural machine translation.
\newblock In {\em Proceedings of the 2015 Conference on Empirical Methods in
  Natural Language Processing, {EMNLP} 2015\/}. pages 1412--1421.

\bibitem[{Ma and Sun(2017)}]{MaSun2017}
Shuming Ma and Xu~Sun. 2017.
\newblock A semantic relevance based neural network for text summarization and
  text simplification.
\newblock {\em CoRR\/} abs/1710.02318.

\bibitem[{Ma et~al.(2017)Ma, Sun, Xu, Wang, Li, and Su}]{srb}
Shuming Ma, Xu~Sun, Jingjing Xu, Houfeng Wang, Wenjie Li, and Qi~Su. 2017.
\newblock Improving semantic relevance for sequence-to-sequence learning of
  chinese social media text summarization.
\newblock In {\em Proceedings of the 55th Annual Meeting of the Association for
  Computational Linguistics, {ACL} 2017, Vancouver, Canada, July 30 - August 4,
  Volume 2: Short Papers\/}. pages 635--640.

\bibitem[{Nallapati et~al.(2016)Nallapati, Zhou, dos Santos,
  G{\"{u}}l{\c{c}}ehre, and Xiang}]{ibmsummarization}
Ramesh Nallapati, Bowen Zhou, C{\'{\i}}cero~Nogueira dos Santos, {\c{C}}aglar
  G{\"{u}}l{\c{c}}ehre, and Bing Xiang. 2016.
\newblock Abstractive text summarization using sequence-to-sequence rnns and
  beyond.
\newblock In {\em Proceedings of the 20th {SIGNLL} Conference on Computational
  Natural Language Learning, CoNLL 2016, Berlin, Germany, August 11-12,
  2016\/}. pages 280--290.

\bibitem[{Narayan and Gardent(2014)}]{NarayanEA2014}
Shashi Narayan and Claire Gardent. 2014.
\newblock Hybrid simplification using deep semantics and machine translation.
\newblock In {\em Proceedings of the 52nd Annual Meeting of the Association for
  Computational Linguistics, {ACL}\/}. pages 435--445.

\bibitem[{Nisioi et~al.(2017)Nisioi, Stajner, Ponzetto, and
  Dinu}]{NisioiEA2017}
Sergiu Nisioi, Sanja Stajner, Simone~Paolo Ponzetto, and Liviu~P. Dinu. 2017.
\newblock Exploring neural text simplification models.
\newblock In {\em Proceedings of the 55th Annual Meeting of the Association for
  Computational Linguistics, {ACL}\/}. pages 85--91.

\bibitem[{Papineni et~al.(2002)Papineni, Roukos, Ward, and Zhu}]{bleu}
Kishore Papineni, Salim Roukos, Todd Ward, and Wei{-}Jing Zhu. 2002.
\newblock Bleu: a method for automatic evaluation of machine translation.
\newblock In {\em Proceedings of the 40th Annual Meeting of the Association for
  Computational Linguistics\/}. pages 311--318.

\bibitem[{Prakash et~al.(2016)Prakash, Hasan, Lee, Datla, Qadir, Liu, and
  Farri}]{paraphrasegeneration}
Aaditya Prakash, Sadid~A. Hasan, Kathy Lee, Vivek~V. Datla, Ashequl Qadir, Joey
  Liu, and Oladimeji Farri. 2016.
\newblock Neural paraphrase generation with stacked residual {LSTM} networks.
\newblock In {\em {COLING} 2016, 26th International Conference on Computational
  Linguistics, Proceedings of the Conference: Technical Papers, December 11-16,
  2016, Osaka, Japan\/}. pages 2923--2934.

\bibitem[{Rush et~al.(2015)Rush, Chopra, and Weston}]{abs}
Alexander~M. Rush, Sumit Chopra, and Jason Weston. 2015.
\newblock A neural attention model for abstractive sentence summarization.
\newblock In {\em Proceedings of the 2015 Conference on Empirical Methods in
  Natural Language Processing, {EMNLP} 2015, Lisbon, Portugal, September 17-21,
  2015\/}. pages 379--389.

\bibitem[{Srivastava et~al.(2014)Srivastava, Hinton, Krizhevsky, Sutskever, and
  Salakhutdinov}]{dropout}
Nitish Srivastava, Geoffrey~E. Hinton, Alex Krizhevsky, Ilya Sutskever, and
  Ruslan Salakhutdinov. 2014.
\newblock Dropout: a simple way to prevent neural networks from overfitting.
\newblock {\em Journal of Machine Learning Research\/} 15(1):1929--1958.

\bibitem[{Sun et~al.(2017{\natexlab{a}})Sun, Ren, Ma, and Wang}]{SunEA2017}
Xu~Sun, Xuancheng Ren, Shuming Ma, and Houfeng Wang. 2017{\natexlab{a}}.
\newblock meprop: Sparsified back propagation for accelerated deep learning
  with reduced overfitting.
\newblock In {\em Proceedings of the 34th International Conference on Machine
  Learning, {ICML} 2017, Sydney, NSW, Australia, 6-11 August 2017\/}. pages
  3299--3308.

\bibitem[{Sun et~al.(2017{\natexlab{b}})Sun, Ren, Ma, Wei, Li, and
  Wang}]{mesimp}
Xu~Sun, Xuancheng Ren, Shuming Ma, Bingzhen Wei, Wei Li, and Houfeng Wang.
  2017{\natexlab{b}}.
\newblock Training simplification and model simplification for deep learning:
  {A} minimal effort back propagation method.
\newblock {\em CoRR\/} abs/1711.06528.

\bibitem[{Sun et~al.(2017{\natexlab{c}})Sun, Wei, Ren, and Ma}]{SunWei2017}
Xu~Sun, Bingzhen Wei, Xuancheng Ren, and Shuming Ma. 2017{\natexlab{c}}.
\newblock Label embedding network: Learning label representation for soft
  training of deep networks.
\newblock {\em CoRR\/} abs/1710.10393.

\bibitem[{Sutskever et~al.(2014)Sutskever, Vinyals, and Le}]{seq2seq}
Ilya Sutskever, Oriol Vinyals, and Quoc~V. Le. 2014.
\newblock Sequence to sequence learning with neural networks.
\newblock In {\em Advances in Neural Information Processing Systems 27: Annual
  Conference on Neural Information Processing Systems 2014\/}. pages
  3104--3112.

\bibitem[{Takase et~al.(2016)Takase, Suzuki, Okazaki, Hirao, and Nagata}]{amr}
Sho Takase, Jun Suzuki, Naoaki Okazaki, Tsutomu Hirao, and Masaaki Nagata.
  2016.
\newblock Neural headline generation on abstract meaning representation.
\newblock In {\em Proceedings of the 2016 Conference on Empirical Methods in
  Natural Language Processing, {EMNLP} 2016, Austin, Texas, USA, November 1-4,
  2016\/}. pages 1054--1059.

\bibitem[{Wang et~al.(2017)Wang, Liu, Sui, and Chang}]{randomwalk}
Kexiang Wang, Tianyu Liu, Zhifang Sui, and Baobao Chang. 2017.
\newblock Affinity-preserving random walk for multi-document summarization.
\newblock In {\em Proceedings of the 2017 Conference on Empirical Methods in
  Natural Language Processing, {EMNLP} 2017, Copenhagen, Denmark, September
  9-11, 2017\/}. pages 210--220.

\bibitem[{Woodsend and Lapata(2011)}]{Woodsend2011}
Kristian Woodsend and Mirella Lapata. 2011.
\newblock Learning to simplify sentences with quasi-synchronous grammar and
  integer programming.
\newblock In {\em Proceedings of the 2011 Conference on Empirical Methods in
  Natural Language Processing, {EMNLP}\/}. pages 409--420.

\bibitem[{Wubben et~al.(2012)Wubben, van~den Bosch, and Krahmer}]{WubbenEA2012}
Sander Wubben, Antal van~den Bosch, and Emiel Krahmer. 2012.
\newblock Sentence simplification by monolingual machine translation.
\newblock In {\em The 50th Annual Meeting of the Association for Computational
  Linguistics, Proceedings of the Conference\/}. pages 1015--1024.

\bibitem[{Xu et~al.(2018)Xu, Sun, Ren, Lin, Wei, and Li}]{dnerre}
Jingjing Xu, Xu~Sun, Xuancheng Ren, Junyang Lin, Binzhen Wei, and Wei Li. 2018.
\newblock Dp-gan: Diversity-promoting generative adversarial network for
  generating informative and diversified text.
\newblock {\em CoRR\/} abs/1802.01345.

\bibitem[{Xu et~al.(2016)Xu, Napoles, Pavlick, Chen, and
  Callison{-}Burch}]{XuEA2016}
Wei Xu, Courtney Napoles, Ellie Pavlick, Quanze Chen, and Chris
  Callison{-}Burch. 2016.
\newblock Optimizing statistical machine translation for text simplification.
\newblock {\em {TACL}\/} 4:401--415.

\bibitem[{Zhang and Lapata(2017)}]{ZhangEA2017}
Xingxing Zhang and Mirella Lapata. 2017.
\newblock Sentence simplification with deep reinforcement learning.
\newblock In {\em Proceedings of the 2017 Conference on Empirical Methods in
  Natural Language Processing, {EMNLP} 2017, Copenhagen, Denmark, September
  9-11, 2017\/}. pages 584--594.

\bibitem[{Zhu et~al.(2010)Zhu, Bernhard, and Gurevych}]{ZhuEA2010}
Zhemin Zhu, Delphine Bernhard, and Iryna Gurevych. 2010.
\newblock A monolingual tree-based translation model for sentence
  simplification.
\newblock In {\em {COLING} 2010\/}. pages 1353--1361.

\end{thebibliography}

\end{document}